\documentclass[10pt,twocolumn,letterpaper]{article}

\usepackage[letterpaper,textwidth=7in,textheight=9in,centering]{geometry}
\usepackage{amsmath}
\usepackage{amssymb}
\usepackage{array}
\usepackage{booktabs}
\usepackage[font=small,labelfont=bf,labelsep=period]{caption}
\usepackage{enumitem}
\usepackage{float}
\usepackage{graphicx}
\usepackage{microtype}
\usepackage{newtxtext,newtxmath}
\usepackage{placeins}
\usepackage{stfloats}
\usepackage{tabularx}
\usepackage{titlesec}
\usepackage{cite}
\usepackage[hidelinks]{hyperref}
\usepackage{xspace}

\titleformat{\section}{\large\bfseries}{\thesection}{0.6em}{}
\titleformat{\subsection}{\normalsize\bfseries}{\thesubsection}{0.6em}{}
\titleformat{\subsubsection}{\normalsize\bfseries}{\thesubsubsection}{0.6em}{}
\titleformat{\paragraph}[runin]{\normalsize\bfseries}{}{0pt}{}
\titlespacing*{\section}{0pt}{2.7ex plus .5ex minus .3ex}{1.2ex plus .2ex}
\titlespacing*{\subsection}{0pt}{2.0ex plus .4ex minus .2ex}{0.8ex plus .2ex}
\titlespacing*{\subsubsection}{0pt}{1.6ex plus .3ex minus .2ex}{0.6ex plus .2ex}
\titlespacing*{\paragraph}{0pt}{1.3ex plus .3ex minus .2ex}{0.6em}

\setlist{nosep,leftmargin=*}
\renewcommand{\arraystretch}{1.10}

\newcommand{\system}{\textsc{SkillEffect}\xspace}
\newcommand{\mib}{\,MiB\xspace}
\newcommand{\code}[1]{\texttt{#1}}
\newcolumntype{Y}{>{\raggedright\arraybackslash}X}
\newcommand{\tablehead}[1]{\multicolumn{1}{c}{\textbf{#1}}}

\title{\Large\bfseries SkillEffect: Checked Lowering for Memory-Bounded Agent Tools}
\author{
  \large\bfseries Yinuo Wang, Yiyu Shi \\[6pt]
  \normalsize\mdseries Department of Computer Science \& Engineering \\[-1pt]
  \normalsize\mdseries University of Notre Dame \\[-1pt]
  \normalsize\mdseries Notre Dame, IN, USA \\[1pt]
  \normalsize\mdseries yinuow800@gmail.com, yshi4@nd.edu
}
\date{}

\hypersetup{
  pdftitle={SkillEffect: Checked Lowering for Memory-Bounded Agent Tools},
  pdfauthor={Yinuo Wang and Yiyu Shi}
}

\begin{document}
\maketitle

\begin{abstract}
Agent Skills can specify procedural and resource obligations for tool use, and
language models instantiate them as concrete programs.  However, when models
turn this guidance into code for existing tool interfaces, even a semantically
correct program may load an entire input and exceed the memory available to one
tool call.  We present
\system, a checked-lowering runtime for computations with a recoverable source
relation, an audited bounded implementation, and a registered output
postcondition.  Before granting execution authority, an independent checker
rebuilds each proposed lowering from the submitted program and immutable input.
Every relation plugin supplies a source recognizer, input-fact extractor,
bounded-IR constructor, arena-bound function, and postcondition; one common
runtime provides checked selection, bounded-VM execution, atomic capacity
leasing, and staged publication.  Generality in \system is architectural rather
than automatic: each supported computation requires an audited relation plugin,
while the dispatch, resource-control, execution, and publication mechanisms are
shared across plugins.  Across six operator families, bounded access
substantially reduces peak memory and improves completion under externally
fixed caps.  Six plugins instantiate the same contract across five execution
patterns, from streaming reduction to bounded-heap Top-\mbox{$k$}.  The XLSX
onboarding study and Top-$k$ extension show that a new relation and a new
retained-state pattern reuse the same trust boundary, while the checker accepts
all evaluated legal configurations and rejects all adversarial proposals.
Together, these results show that one checked-lowering architecture can enforce
heterogeneous registered memory relations at Agent tool dispatch.
\end{abstract}

\section{Introduction}

Reusable Agent Skills increasingly supply procedural knowledge to language
models: which tool to call, which fields matter, how large inputs should be
processed, and how results should be checked
\cite{chen2026skvm,ouyang2026skcc,tan2026skillzip}.  The same procedure can compile
into radically different physical executions.  A plan that calls
\code{read\_csv} materializes a table; a lazy scan can project, filter, and
aggregate it in bounded space.  Loading an entire scientific matrix is different
from opening only its metadata, even when both answer the same query.  Under a
server-sized budget this difference is a performance issue.  Under a 100\mib
tool-stage cap, it can determine whether the tool call completes.  A Skill
recommendation alone is insufficient: in our external smolagents harness, all
16 otherwise valid tool calls choose the eager access mode when no resource
feedback is provided.

This choice has a direct serving consequence.  Agent platforms execute many
tool calls in isolated sandboxes, and the per-invocation memory reservation
sets both failure isolation and the number of sessions a host can sustain.
Tight allocations are already ordinary infrastructure settings rather than a
paper-specific thought experiment: Modal Functions and Sandboxes request
128\mib by default and can be given an explicit hard limit; E2B offers sandbox
memory configurations beginning at 512\mib, and paid plans permit CPU/RAM
customization; Cloudflare Workers limit each isolate to
128 MB; and AWS Lambda's default and minimum setting is 128 MB
\cite{modalresources,e2bsandbox,cloudflarelimits,awslambdamemory}.
Daytona's fleet model uses one sandbox per user, task, or agent; resources are
reserved per sandbox and the running fleet draws from an organization-level
compute pool
\cite{daytonascale}.  Together, these offerings establish 128--512 MB-class
allocations and limits as an ordinary sandbox/serverless capacity region.
Reserving for an eager worst case wastes that pool; admitting
an optimistic plan risks killing a trajectory after the model has already
spent tokens and latency.  The useful control point is therefore the pending
dispatch: after the model has made its plan concrete, but before the tool
process consumes the tenant's allocation.

\begin{figure*}[t]
\centering
\includegraphics[width=\textwidth]{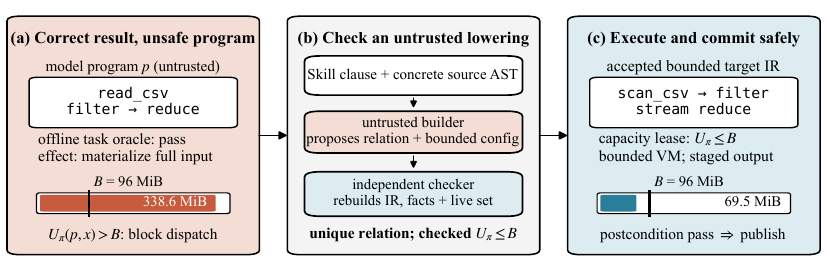}
\caption{The \system trust boundary, illustrated with the two-million-row
Polars S2 relation.  The 338.6/69.5\mib peaks are paired generous-cap operator
measurements; 96\mib is the same-input development cap, not an external-sweep
tier.  The builder's lowering is untrusted.  An independent checker identifies
the unique matching relation, rebuilds source semantics and input facts, and
validates the target configuration and live set; a bounded VM executes
only the accepted IR under a capacity lease and publishes staged output only
after its registered postcondition passes.}
\label{fig:idea}
\end{figure*}

Existing layers leave a concrete systems gap.  Skill compression can preserve
procedural contracts or measure whether control obligations survive
compression, but does not generally assign a physical live-set effect to the
program selected by the model
\cite{tan2026skillzip,hou2026control}.  Cost-aware planners use profiles to
prefer cheaper tools \cite{wu2025catp}, but a profile does not construct a
relation-preserving bounded program.  A plan gate or cgroup can reject or stop
an unsafe process \cite{zhang2026agentlibos,zheng2026agentcgroup}, but cannot
recover the intended result.  Conversely, replacing \code{pandas} with a
database or streaming operator is not correct merely because it uses less
memory: filters, output keys, reduction semantics, and edge cases must still
match the declared task \cite{baziotis2024dias,pan2026rulescript,knoth2019resyn}.

Skills are heterogeneous by construction: a short API wrapper, a tabular
analysis procedure, and a scientific-file workflow need not share the same
resource bottleneck.  We target the important class for which the Skill states a resource-sensitive access
obligation, the generated plan exposes a recoverable declarative computation,
a bounded implementation exists, and the declared result admits a registered
postcondition.  CSV aggregation, sequence reduction, scientific metadata,
chunked arrays, workbook streaming, and risk-ranked audit-log selection
instantiate different members of this class.

Across these members, the reusable unit is a checked relation
$r=\langle S_r,I_r,L_r,A_r,G_r\rangle$.  Here $S_r$ recognizes the source
computation, $I_r$ inspects immutable input facts, $L_r$ constructs canonical
bounded IR, $A_r$ derives its arena, and $G_r$ checks the result before
publication.
Skill routing establishes the task and tool context, after which the independent
checker reconstructs and validates a concrete relation instance before dispatch.
A new computation therefore extends \system by implementing these five
relation-specific obligations while reusing the common dispatch trust boundary.

This paper asks the following question: can an agent runtime turn an over-budget
model-generated tool plan into a checked bounded implementation before dispatch,
without changing the declared task result?  The evaluation has two complementary
roles.  Operator characterization measures
the physical opportunity across six families: a concrete access choice can
change the live set without changing the declared answer.  The \system
experiments then instantiate one relation contract across six plugins and
exercise the complete proposal--check--execute boundary, from independent
reconstruction through bounded-VM execution, capacity leasing, and staged
publication.  The extension studies use XLSX to measure onboarding cost and
Top-$k$ selection to test a new bounded-state pattern under the same contract.

The contributions are:

\begin{itemize}
  \item a physical characterization of a concrete agent-systems failure mode
  across six tool families: eager and bounded access paths can preserve the
  same declared result while producing different live sets and hard-cap
  completion outcomes, including correct model-generated plans that fail when
  executed eagerly;
  \item a proposal--check--execute trust boundary built around a common
  five-obligation relation contract in which semantics and resource arithmetic
  are relation-specific, while binding, unique selection, admission, bounded
  execution, leasing, and publication are shared; and
  \item an evaluation of six plugins across five bounded-execution patterns,
  including (i) held-out XLSX onboarding to measure extension cost for an
  existing pattern and (ii) Top-$k$ selection to test a qualitatively new
  retained-state pattern without changing the generic core.
\end{itemize}

Figure~\ref{fig:idea} instantiates the proposal--check--execute boundary for a
concrete Polars program generated for a routed Skill;
Section~\ref{sec:design} generalizes the same relation interface across the
registry.

\section{System Model and Guarantees}

\subsection{Resource and Admission Model}

\system implements a guarded optimization boundary for memory-isolated tool
serving.  An accepted dispatch carries both a task-preserving implementation
and a capacity commitment.  When neither the original plan nor an audited
lowering can establish those conditions, control returns before tool
allocation so the scheduler can choose another tool, budget, or execution
site.

The selected Skill describes the intended tool procedure and its resource
requirements.  At dispatch, \system matches the submitted program against the
audited relations in its registry.  A matching relation binds the recognized
source computation and immutable input facts to a bounded target, a memory
bound, and a publication check.  The LLM and its GPU memory are outside this
tool-stage capacity domain.  Let $p$ be the generated but uncommitted tool
program, $x$ its immutable input, and $\pi$ the platform manifest.  We model
the peak physical memory of a captured plan as

\begin{equation}
\begin{aligned}
W_e(x)&=\begin{cases}
n_e(x),&\text{eager},\\[-2pt]
\min\!\left(n_e(x),\ell_e\right),&\text{bounded},
\end{cases}\\[-1pt]
U_{\pi}(p,x)&=F_{\pi}+M_{\pi}
+\max_t\sum_{e\in\mathcal{A}_p(t)}
\bigl[\alpha_{\pi,e}W_e(x)+R_e(x)\bigr].
\end{aligned}
\label{eq:memory}
\end{equation}

$\mathcal{A}_p(t)$ is the set of effects simultaneously live at time $t$;
$n_e(x)$ is the input-sized state of effect $e$, and $\ell_e$ is its bounded
window.  Thus $W_e(x)$ retains the full state for an eager effect and at most
$\ell_e$ for a bounded one.  $F_{\pi}$ is fixed runtime state, $M_{\pi}$ a
frozen platform reserve, $\alpha_{\pi,e}$ maps logical live state to physical
occupancy, and $R_e$ covers scratch and output state.  The maximum over $t$
composes overlapping effects while allowing sequential phases to reuse capacity.  Operator
characterization instantiates these terms from calibrated fresh-cgroup envelopes.  Across the
six registered plugins, the checker instead derives the variable
arena and staged-output terms directly and adds hash-bound runtime and I/O
reserves from the platform manifest.  The assurance boundary combines a
source-derived variable arena with empirical constants for interpreter,
allocator, and kernel state.

\paragraph{Checked relations and rule selection.}
Let $\mathcal{R}$ be the six-relation registry, $B$ the tool-stage capacity
budget, $G_r$ the online postcondition registered with relation $r$, and $V_q$
the exact verifier for task $q$ used in evaluation.  For each applicable
$r\in\mathcal{R}$, an untrusted builder may propose a checked record $z_r$
containing a bounded configuration and target IR.  The independent checker
evaluates $C_r(z_r;p,x,\pi)$ by
reconstructing the complete allowed source AST, its source-derived semantic
parameters, immutable input facts, and the rule's arithmetic and dialect
obligations.  It also validates the proposal's bounded window and staged-output
capacities and rebuilds the canonical target IR and live-set bound for that
configuration.  This closed source-to-target
relation establishes source-to-target admissibility; $G_r$ checks the online
result, while $V_q$ scores the exact task result in the evaluation.

The online postcondition is part of the registered relation.  Rule selection
and semantic parameters are reconstructed from the submitted source and input;
window and staged-output capacities remain checked proposal parameters.  None
depends on the evaluation oracle's expected values.  Define the accepted set

\begin{equation}
\mathcal{C}_{B,\pi}(p,x)=
\left\{(r,z_r)\ \middle|\
\begin{array}{l}
r\in\mathcal{R},\ C_r(z_r;p,x,\pi)=\mathrm{pass},\\
z_r.U\leq B
\end{array}
\right\}.
\label{eq:candidates}
\end{equation}
The registry implements
\begin{equation}
(r^\star,z^\star)=
\begin{cases}
c,&\mathcal{C}_{B,\pi}(p,x)=\{c\},\\
\bot,&\text{otherwise}.
\end{cases}
\label{eq:select}
\end{equation}

When selection succeeds, $z^\star.T$ is the bounded IR supplied to the VM.
Zero accepted rules cause abstention, while multiple accepted rules cause an
ambiguity rejection.  The certifier receives neither a task nor a family
label.  The operator-characterization frontend has a separate recognized-identity
case and ordered typed or guarded lowerings; it provides breadth evidence but not the
independent checked relation of Equations~\ref{eq:candidates}
and~\ref{eq:select}.

\paragraph{Admission policy.}
Admission and abstention have asymmetric consequences under a hard cap.
Dispatching an unjustified plan may terminate tool execution and discard the
trajectory, whereas abstention occurs before bounded-VM execution or
publication and returns control to the caller.  Unknown effects and ambiguous
rule matches return control before capacity is consumed.  Admission certifies
feasibility under the current registry and cap, while abstention leaves
alternative tools, budgets, or execution sites available.

\paragraph{Capacity and publication.}
For an admitted request $i$, let $r_i$ and $z_i^\star$ be its selected relation
and record, $y_i$ its staged result, and $U_i=z_i^\star.U$ its checked peak.
Let $H$ be the configured cgroup \code{memory.max}, $P_i$ the measured peak of
an exclusive request cgroup, and $A_t$ the active leases at publication time
$t$.  Let $\mathbf{E}_{\mathrm{limit},i}$ be the applicable cgroup's
\code{max}/\code{oom}/\code{oom\_kill}/\code{oom\_group\_kill} event vector,
and let $\sigma_i$ be its observed swap bytes (for the request's exclusive
cgroup or the common shared cgroup).  Write $\mathsf{publish}_i$ for the event
that $y_i$ becomes externally visible.
Before acquiring a lease, the runtime requires

\begin{equation}
U_i\leq H\leq B.
\label{eq:cgroup-bound}
\end{equation}

The SQLite ledger atomically admits a lease only when the sum of bytes in
\code{reserved}, \code{running}, and \code{verified} states remains at most
$B$.  In the shared-capacity experiments, ledger capacity and cgroup limit are
configured to the same value, $B=H$.  Output publication satisfies

\begin{equation}
\begin{aligned}
\mathsf{publish}_i\Longrightarrow{}&
G_{r_i}(y_i)=\mathrm{pass}\\[-1pt]
&\land\ \mathbf{E}_{\mathrm{limit},i}=\mathbf{0}
\ \land\ \sigma_i=0\\[-1pt]
&\land\
\begin{cases}
P_i\leq U_i,&\text{exclusive},\\
\sum_{j\in A_t}U_j\leq B,&\text{shared}.
\end{cases}
\end{aligned}
\label{eq:commit}
\end{equation}

The exclusive path compares a fresh request cgroup's peak with that request's
checked bound.  The shared path uses the atomic sum of active leases as its
admission invariant, while the common cgroup and zero limit events guard the
capacity domain.  Formal physical runs begin with zero event counters.  The VM
result and serialized staged file must both satisfy the registered online
postcondition before no-overwrite atomic publication.  In the oracle-isolation
experiments, $V_q$ is applied by the host only after the committed container
exits and therefore supplies evaluation rather than execution authority.

\subsection{Threat Model}

Our threat model distrusts Skill text, generated Python, the proposed target
IR, semantic and resource witnesses, and all proposal hashes, whether they are
malformed accidentally or crafted maliciously.  A malicious proposer may recompute those
hashes after changing a proposal.  Separately, a worker may crash at any
lifecycle boundary.  The trusted base comprises the checker, bounded VM,
lease/commit runtime, hash-bound platform manifest, immutable input, registered
online postcondition, container image, and kernel cgroup mechanism; the proposer cannot
modify them.  The present system covers deterministic, local, read-only tools.
Irreversible remote effects such as payment or email require an effect
transaction protocol.

The CPython parser and each rule-specific input inspector are part of the TCB.
The current checker accepts a closed source grammar, but it does not yet place
an independently enforced byte, AST-node, depth, or certification-time limit
around malicious source.  Production deployment must isolate preflight and
bound these quantities before parsing; our adversarial mutations test semantic
and binding confusion, not parser denial of service.

We charge the entire tool cgroup: interpreter, anonymous memory, file-backed
pages and page cache visible to its cgroup, libraries, and result buffers.  The
cap applies to local tool-stage physical memory; model GPU memory is outside
this domain.  The six-family operator characterization uses calibrated,
platform-specific envelopes to measure physical leverage.  In \system, the
checker derives the variable live set while the platform manifest
supplies empirical interpreter, allocator, I/O, and kernel reserves.

Input-fact reconstruction happens during preflight in a separate fresh cgroup
with the same hard cap and swap disabled; the online oracle-isolation protocol
charges that scan, including its page cache, as its own capped transaction.
The resulting facts and input hash are then immutable certificate inputs to the
execution transaction and are frozen before lease admission.  Fact extraction
is format dependent: CSV/FASTA/FCS inspections are
streaming or header bounded, whereas the current Zarr checker scans the raw
input to derive an exact reduction.  We report certification and execution as
separate capped phases, with fact-extraction cost charged to certification.

\subsection{Trust Contract and Capacity Invariant}

\paragraph{Semantic TCB contract.}
For each registered relation, the checker and bounded VM are trusted to
implement the audited source observation and bounded target.  Acceptance
selects that implementation and validates its proposal parameters; $G_r$ checks
the online output contract, while $V_q$ supplies exact experimental evaluation.
The mutation study in Section~\ref{sec:checker-eval} tests this implementation
boundary rather than treating it as a machine-checked proof.

\paragraph{Lease accounting invariant.}
Let $A_t$ be the active leases after any ledger
transition.  Initially $A_0=\emptyset$; acquire commits only when
$\sum_{j\in A_t}U_j+U_i\leq B$, lifecycle transitions preserve a lease's
charge, and release removes it.  Induction over serialized transitions gives
$\sum_{j\in A_t}U_j\leq B$ at every reachable state, including intermediate
lifecycle states.
If every admitted $U_i$ bounds the physical live set charged to its cgroup with
swap disabled, physical occupancy for admitted execution transactions is at
most $B$ as a corollary; the cgroup independently enforces the cap.  The ledger
invariant covers execution transactions.  A multi-tenant deployment extends the
same host-level ledger to concurrent preflight transactions.

\section{Design}
\label{sec:design}

\system follows a three-stage proposal--check--execute trust boundary.  First,
the trusted path reparses the complete source and recovers its concrete
resource effect.  Second, a builder proposes a bounded target, while an
independent checker reconstructs the admitted relation, target IR, and live-set
bound.  Third, the runtime acquires capacity, executes only checker-rebuilt IR
in the bounded VM, and gates publication on the registered postcondition.  A
builder may be an LLM, a heuristic frontend, or a conventional optimizer; its
target, witnesses, and bindings remain proposals, so unsupported or incorrect
constructions become abstentions.  The six-family operator characterization
uses closed lowering schemas and calibrated envelopes to establish the
available physical leverage; the registered runtime supplies the independent
checker, bounded VM, platform manifest, capacity lease, online postconditions,
and cgroup enforcement.

\subsection{Program Analysis and Relation Recovery}

The frontend parses the complete model-generated Python AST and records
resource-relevant calls and keyword arguments: \code{read\_only}, \code{usecols},
\code{only\_text}, \code{backed}, array slicing, lazy scans, streaming collect,
and iterator materialization.  All semantics-bearing parameters must be
literal or reconstructed from input facts by the trusted checker.  Dynamic
calls, unresolved names, and ambiguous parameters remain unknown.  This
prevents a common failure of prompt-only guards: preserving the sentence ``use
streaming for large files'' does not establish that the generated program
streams.

For each relevant operator, the core IR records
$e=\langle f,o,m,\ell,b\rangle$, where $f$ is the tool family, $o$ the
operator, $m$ the access mode, $\ell$ the live-set description, and $b$ the
bound kind.  Source lines, evidence, and artifact hashes are retained as
provenance, while admission depends on the reconstructed checked relation.
Access modes include full materialization, projected materialization,
metadata-only or backed access, chunked reduction, streaming collection, and
single-record iteration.  Unknown calls remain \emph{unproven}.

The IR separates an operator's semantic live window from its physical bound.
For example, ``one Zarr chunk plus a fixed histogram'' follows from the
operator, while the physical number also includes the interpreter, allocator,
libraries, and page cache.  In operator characterization,
Equation~\ref{eq:memory} is instantiated from the concrete input and a
hash-bound platform manifest; $F_{\pi}$, $\alpha_{\pi,e}$, and $M_{\pi}$ are
calibrated from fresh-cgroup runs.  In the registered runtime, the independent
checker derives the live window from the accepted source grammar and adds a
platform reserve for fixed interpreter and kernel state.  Only a
checker-accepted bounded mode can authorize small-cap dispatch; eager estimates
remain diagnostic.

\subsection{Independent Lowering Check}

A registered relation is an executable implementation of the common contract
$r=\langle S_r,I_r,L_r,A_r,G_r\rangle$.  A proposed record $z$ carries a
bounded configuration $\rho=z.\rho=(w,s)\in D_r$, consisting of an input
window and staged-output capacity.  Relation $r$ recomputes
\[
\begin{aligned}
\theta_r &= S_r(\operatorname{AST}(p)), &
f_r &= I_r(x),\\
\widehat T_r &= L_r(\theta_r,f_r;\rho), &
\widehat U_r &= R^{\mathrm{run}}_\pi+R^{\mathrm{io}}_\pi
                +A_r(\widehat T_r,f_r)+s .
\end{aligned}
\]
Here $D_r$ is the relation's admissible configuration domain, $\theta_r$ the
recovered source semantics, $f_r$ the immutable input facts, $\widehat T_r$ the
checker-rebuilt target IR, and $\widehat U_r$ its checked peak.  $S_r$ reparses
the complete source under a closed grammar, $I_r$ inspects the immutable input,
$L_r$ constructs canonical bounded-VM IR, $A_r$ computes the relation-specific
arena, and $G_r$ checks the result before publication.  The hash-bound reserves
$R^{\mathrm{run}}_\pi$ and $R^{\mathrm{io}}_\pi$ instantiate the fixed platform
component of Equation~\ref{eq:memory}.  Thus $\widehat U_r$ is the
relation-specific instance of $U_\pi(z.T,x)$; after unique selection, it becomes
the request bound $U_i$.  The implemented acceptance predicate is
\[
\begin{aligned}
&C_r(z;p,x,\pi)=\mathrm{pass}\ \Longleftrightarrow\\
&\quad \theta_r\ne\bot\ \land\ \rho\in D_r\\
&\quad {}\land\ z.\theta=\theta_r\ \land\ z.f=f_r\\
&\quad {}\land\ z.T=\widehat T_r\ \land\ z.U=\widehat U_r\\
&\quad {}\land\ \operatorname{Bind}_\pi(z,p,x).
\end{aligned}
\]
$\operatorname{Bind}_\pi$ checks the source, input, platform manifest, and
trusted-code hashes.  Equality is structural over the canonical JSON records;
the checker derives every right-hand-side value independently of the builder.
Concretely, each relation implements five obligations: recognize the complete
source program with $S_r$, reconstruct immutable input facts with $I_r$,
construct the bounded operator sequence with $L_r$, derive its arena with
$A_r$, and check the publication predicate with $G_r$.  The runtime supplies
the common binding, unique selection, capacity lease, bounded-VM execution,
and staged-publication path.

This contract deliberately does not synthesize arbitrary program equivalences.
Relation authors audit the source language accepted by $S_r$, the semantic facts
recovered by $I_r$, the target semantics constructed by $L_r$, the arena
arithmetic in $A_r$, and the publication predicate $G_r$.  \system's reusable
contribution is to make such audited relations composable with one common
dispatch and resource-control path.

When a submitted AST matches a supported source schema and its checked effect
exceeds $B$, the builder extracts the literals required by a registered rule
and proposes $\rho$ and a target record.  Dynamic calls, unresolved names,
nonliteral predicates, ambiguous output shapes, unsupported expressions, or
missing output keys make $S_r$ return $\bot$.  Table~\ref{tab:ops} shows how
the six executable relations instantiate the same five obligations.  The first
five cover four bounded-execution patterns: streaming relational pipelines
(CSV), iterator/row-window reductions (FASTA and XLSX), metadata projection
(FCS), and chunked reduction (Zarr).  Top-$k$ adds an explicit bounded-state
pattern whose retained state grows with $k$, not with input cardinality.
These patterns demonstrate diversity in bounded-state structure and access
behavior; they are not intended to cover arbitrary Python programs or tool
APIs.
Appendix~\ref{app:operators} separately characterizes eager and bounded access
modes across six operator families.

\begin{table*}[t]
\centering
\captionsetup{width=0.98\textwidth}
\caption{Six executable plugins for one checked-relation contract, spanning
five bounded-execution patterns.  Each row supplies $S_r$, $I_r$, $L_r$, $A_r$,
and $G_r$; the common checker additionally validates source, input, platform,
target-IR, and bound bindings.  ``Retained state'' describes the
relation-specific variable state rather than the fixed platform reserve.  The
table separates relation-specific obligations from the common runtime
mechanisms, which are not reimplemented per plugin.}
\label{tab:ops}
\small
\begingroup
\setlength{\tabcolsep}{2.8pt}
\renewcommand{\arraystretch}{1.10}
\renewcommand{\tabularxcolumn}[1]{m{#1}}
\begin{tabularx}{0.98\textwidth}{@{}
  >{\centering\arraybackslash}m{0.095\textwidth}
  >{\centering\arraybackslash}m{0.175\textwidth}
  >{\centering\arraybackslash}m{0.225\textwidth}
  >{\centering\arraybackslash}m{0.175\textwidth}
  >{\centering\arraybackslash}X@{}}
\toprule
\tablehead{Relation} & \tablehead{Source pattern} &
\tablehead{Execution pattern and retained state} & \tablehead{Input facts $I_r$} &
\tablehead{Postcondition $G_r$} \\
\midrule
CSV / Polars & Eager relational pipeline & Streaming relational pipeline:
projection, filters, and
integer reduction; input window plus constant aggregates & Header and strict
stream scan & Schema/range and aggregate invariants \\
FASTA / Biopython & Sequence-list materialization & Iterator/record reduction;
input window plus fixed histogram & Bounded-line sequence scan & Record/base
cardinality and conservation \\
FCS / FlowIO & Event-payload loading & Metadata projection from header/TEXT;
bounded metadata state & File offsets, header, and TEXT segment & Requested
metadata reconstructed from input facts \\
Zarr & Full-array materialization & Chunked incremental reduction;
one chunk plus fixed histogram & Array metadata and raw chunks & Shape, unsigned
sum, and histogram reconstructed from chunks \\
XLSX & Worksheet materialization & Iterator/row-window reduction; row window
plus constant aggregates & Workbook metadata and row scan & Three aggregates
reconstructed from row facts \\
Audit Top-$k$ & Global materialization and sort & Bounded-heap Top-$k$; input
window plus $k$ records & Strict JSONL stream scan & Cardinality, unique IDs,
value ranges, and registered ordering \\
\bottomrule
\end{tabularx}
\endgroup
\end{table*}

Registry selection therefore dispatches a concrete checker implementation.
The bounded VM receives only $\widehat T_r$, and publication occurs only after
$G_r$ accepts its result.  Operator characterization uses a task verifier after
execution; the registered runtime performs the source, target, resource, and
publication checks shown above.  The builder that proposes a target may change
without entering this trusted path; authority is granted only to the concrete
record accepted by the relation checker.

\subsection{Capacity-Safe Execution and Publication}

The runtime applies the following decision procedure:

\begin{enumerate}
  \item reconstruct the complete source plan, concrete input facts, and every
  supported identity or lowering candidate;
  \item admit an operator-characterization identity or a unique
  checker-accepted target only when its bound fits $B$, then acquire capacity
  and execute the bounded handler or target IR into staged output under the
  cgroup cap; and
  \item publish only when the registered online postcondition and the physical resource invariant
  pass; otherwise discard staged output and fail closed.
\end{enumerate}

The cgroup enforces $B$ and records
\code{memory.max}, \code{memory.peak}, \code{memory.events}, swap, process exit,
and the container's hard-limit state.  The analytical bound governs admission,
while these counters provide independent post-execution physical evidence.

\FloatBarrier
\section{Evaluation}

We organize the evaluation around three questions:

\begin{enumerate}
  \item When do resource-sensitive access modes change completion under
  externally fixed memory caps?
  \item Can one checked-relation contract recover bounded executions from
  semantically correct model-generated plans, onboard a relation withheld from
  the initial registry without changing the core trust boundary, and support a
  different bounded-state pattern?
  \item Can the trust boundary reject invalid proposals, safely publish
  accepted executions, and operate through Agent harnesses under concurrency
  and failures?
\end{enumerate}

The studies follow the same three layers as the contributions.  Physical
characterization uses operator scaling, external caps, natural plans, latency,
and construction baselines to establish the resource effect and connect it to
generated programs.  Checked-runtime studies combine legal and mutated
configurations, relation onboarding, the Top-$k$ bounded-state extension,
unique rule selection, and oracle isolation to exercise independent
reconstruction, bounded execution, and publication.  Capacity and Agent
studies cover harness dispatch, lease competition, and lifecycle failures.
The six-family study measures physical breadth; six plugins instantiate the
common contract and exercise checker/runtime reuse.  Five workloads occur in
both sets, with AnnData unique to operator characterization and audit-log
Top-$k$ unique to the checked registry.  Detailed operator scaling,
development-cap prompting and repair, frontend coverage, latency, and
per-relation physical accounting appear in the appendix.

\subsection{Experimental Methodology}

\paragraph{Workloads.}
We use six tool families and four parameterizations per family,
forming 24 task--Skill pairs.  Each task has a deterministic generator and
verifier.  Inputs scale across three sizes for the operator study; the largest
contains 100K spreadsheet rows, 2M cytometry events, a $150\mathrm{K}\times512$
AnnData matrix, a $65{,}536\times1{,}024$ Zarr array, 2M CSV rows, or 500K FASTA
records.  Together they form a controlled benchmark of six distinct
resource-sensitive operator families.

\paragraph{Capacity protocols.}
We use two capacity protocols.  The mechanism matrix uses one development cap
per family, placed between the measured bounded and eager peaks to exercise
construction and enforcement.  A separate externally fixed cap sweep fixes
64, 128, 256, 512, 1024, and 2048\mib before any sweep outcome.  The
sweep reports completion over all tasks, semantic qualification at the largest
tier, and the shared-qualified subset separately; a baseline's invalid program
is reported as a semantic failure rather than a memory-system loss.  The natural-plan mechanism study
reuses the development caps unchanged.  We run every physical cell three times
in balanced order.
A run succeeds only when the process exits normally, the verifier passes, the
cgroup measurement is valid, swap remains zero, and no OOM or OOM-kill event
occurs.  We record nonfatal \code{memory.events.max} pressure rather than
misclassifying a successful reclaim-bound execution as a failure.
This rule applies to operator-characterization and baseline cells.  \system
publication applies the stricter Equation~\ref{eq:commit} gate and requires all
four limit-event counters to remain zero.
Repetitions establish stability and are not counted as independent tasks.

\paragraph{Testbed.}
Physical tool executions use the pinned \code{skilleffect-tools:v1} image under
Docker Engine via Colima on an Apple M5 MacBook Air (10 CPU cores, 16\,GiB RAM).
The Linux 6.8/aarch64 VM exposes four vCPUs and cgroup v2; each formal container
receives two vCPUs, a 256-process limit, its stated \code{memory.max}, and no
swap allowance.  \system uses CPython 3.11.15 and hash-binds its
runtime, checker, VM, and platform manifest.  Model generation runs separately
with Qwen2.5-14B-Instruct-AWQ or Mistral-7B served on one A800-80GB PCIe GPU;
GPU memory is outside the tool cgroup cap.  The uncapped throughput control
runs on that server with 112 exposed vCPUs and approximately 979.5\,GiB
available host RAM; it contributes no physical-memory result.

\paragraph{Baselines.}
We compare three kinds of control.  \emph{Constructive} baselines are an
always-pinned resource-clause prompt, ordinary retry with resource feedback,
and the \emph{implemented composition}, which chooses the first statically
bounded plan from always-pin, retry, full-Skill, and compressed-Skill
generations.  \emph{Reject-only} controls are a profile gate and a concrete
plan gate plus cgroup; they measure safe abstention and are not presented as
program constructors.  Direct eager execution is the unmodified reference.
Every executable arm is enforced under the same cap and uses no task-verifier
oracle during construction.
Appendix~\ref{app:protocols} specifies the cap-feedback, natural-plan, agent,
bounded-only, oracle-isolation, and matched-latency protocols.  Each uses frozen
inputs and attempt budgets; repeated executions measure stability rather than
adding independent tasks.

\subsection{Hard-Cap Completion}
\label{sec:external-cap}

The operator study in Appendix~\ref{app:operators} establishes the underlying
physical leverage: across all six families, bounded and eager implementations
produce the same verified results while their largest-input peaks differ by
$3.75$--$24.25\times$ (median $8.45\times$).  We now evaluate whether that
leverage changes completion under caps fixed independently of those peaks.  In
this study, ``bounded lowering'' names the operator-characterization reference;
the \system trust-boundary experiments are reported separately.

\begin{figure}[t]
\centering
\includegraphics[width=\columnwidth]{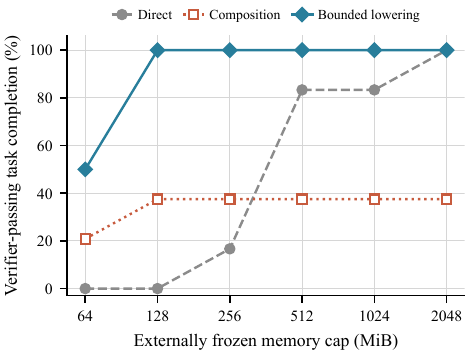}
\caption{Verifier-passing task completion under six externally frozen binary
caps across all 24 tasks.  Markers are the measured quota tiers; connecting
segments aid comparison and do not denote intermediate-cap observations.  The
composition series includes its semantic coverage limit.  On the same nine tasks for which the composition is
valid, it and the bounded-lowering arm have identical completion at every cap,
as quantified in the preceding paragraph.}
\label{fig:external-cap}
\end{figure}

The preregistered sweep contains 1,296 task--cap--repeat--arm cells: 1,116
physical executions and 180 composition abstentions fixed by the absence of a
candidate plan.  All cells are present; every physical measurement is valid,
swap is zero, and there is no monotonicity violation.  Among 750 successful
physical executions, 195 record nonzero \code{memory.events.max} pressure but
none records an OOM event; these are valid boundary completions, not zero-pressure
runs.  Figure~\ref{fig:external-cap}
reports the result without choosing caps from observed peaks.

Across all 24 tasks, direct execution completes 0, 0, 4, 20, 20, and 24 tasks
at 64--2048\mib.  The bounded-lowering arm completes 12 tasks at 64\mib and all
24 from 128\mib onward.  Its normalized log-cap AUC is 0.950, versus 0.467 for direct
execution; its stable C50/C90 are $\leq64$/128\mib, versus 512/2048\mib.  Every one
of the six families lowers its median stable threshold by at least one complete
binary tier.  The 12 failures at 64\mib are all physical-cap failures rather
than abstentions: the sweep deliberately executes each frozen candidate at
every tier, and all 36 repeats for the four AnnData, four Polars, and four Zarr
tasks are OOM-killed, with zero verifier mismatch.  At 128\mib every lowered
task succeeds.

We define normalized log-cap AUC as the trapezoidal area under completion as a
function of $\log_2 B$, divided by the tested log-cap width.  A stable C$x$ is
the smallest tested tier whose task-completion rate is at least $x$ and remains
at least $x$ at every larger tier; each task--tier success additionally requires
all three repeats to pass.  Because 64\mib is the smallest tested tier,
the bounded-lowering arm's C50 is left-censored and should be read as $\leq64\mib$.

The largest gains occur below 512\mib.  Relative to direct execution, lowering
adds 12, 24, and 20 completed tasks at 64, 128, and 256\mib.  At 512 and
1024\mib the remaining four-task gain comes entirely from Zarr; at 2048\mib
both paths complete all 24 tasks.  The sweep therefore locates both the tight-cap
operating region and the point at which additional lowering capacity ceases to
change completion.

The implemented composition is semantically valid at 2\,GiB for only 9/24
tasks, so its all-task AUC of 0.358 combines coverage and memory feasibility.
We therefore make the stronger comparison on exactly those nine tasks.  Both
the composition and bounded-lowering arm complete 5/9 at 64\mib and 9/9 at every larger
tier, giving identical shared-subset AUCs of 0.956.  Bounded lowering therefore
matches the composition whenever it already supplies a correct bounded plan.  Its
additional benefit comes from constructing verifier-valid bounded plans for
the other fifteen tasks.

The development-cap construction study reaches the same qualitative boundary:
always-pin and feedback-based retry construct verifier-valid bounded programs
for 9/24 and 4/24 tasks, respectively.  Appendix~\ref{app:development} reports
the full construction and failure breakdown.

\subsection{Recovering Bounded Execution from Model Plans}

We next connect operator characterization to model-generated code.  The frozen
generation pool contains 72 plans.  Twenty-five attempts pass the
verifier, covering 10 of 24 tasks.  The connection experiment therefore
contains those 10 tasks; the model did not produce a correct plan for the
other 14 within the frozen attempt budget.  Of the selected plans, five are already bounded:
one AnnData task and four Biopython tasks.  Five are correct but unsafe: one
FlowIO plan materializes event data and four Polars plans eagerly load the CSV.

The bounded lowerer leaves all five bounded plans unchanged.  For each unsafe plan, it
recovers its fields or predicates, constructs a bounded target, and re-verifies
the result.  Under the same frozen family caps, unchanged natural plans
complete only the five already-safe tasks (15/30 repeated runs) and incur 15
cap failures.  The bounded lowerer completes all 10 tasks and 30/30 runs with no cap
failure.  Among the five unsafe tasks, the comparison is 0/5 versus 5/5.

Median successful peaks for the lowered FlowIO plan and four Polars plans are
41.66, 81.22, 78.72, 78.73, and 80.48\mib, respectively.  Across the five safe
no-op tasks, the median peaks of the direct and bounded-lowering arms differ by at most
0.21\mib, confirming that an available lowerer does not itself trigger rewriting.

The Mistral-7B replication yields only four verifier-valid attempts across two
tasks, versus 25 attempts across 10 tasks for Qwen2.5-14B.  This exposes a
semantic capability floor before memory control becomes relevant.

\subsection{Validating Relation Proposals}
\label{sec:checker-eval}

We first evaluate checker discrimination on a frozen five-relation configuration
shared by the main admission study; the later Top-$k$ extension is evaluated
under its own separately frozen legal/adversarial schedule in
Section~\ref{sec:relation-onboarding}.  From one accepted record per relation,
we construct a two-sided suite.  The positive side varies legal VM chunk and
staged-output capacities, producing 60 valid configurations.  The mutation side
changes target or source semantics, resource arithmetic, immutable-input facts,
the output gate, platform bindings, or source--target consistency, producing
500 distinct proposals.  Every mutation includes recomputed outer and nested
hashes, so the decision depends on the reconstructed relation rather than a
stale checksum.

The checker independently reconstructs source semantics, input facts, the
canonical target IR, and its live-set bound.  It accepts all 60 legal
configurations and rejects all 500 mutations.  Table~\ref{tab:mutation}
decomposes these decisions by the property changed.

\begin{table}[t]
\centering
\caption{Checker decisions in the frozen five-relation validation matrix;
Top-$k$ extension checks are reported separately in
Section~\ref{sec:relation-onboarding}.  Legal configurations vary VM resources;
mutations alter semantics, resources, or bindings after recomputing all affected
hashes.}
\label{tab:mutation}
\small
\begin{tabular}{@{}cc@{}}
\toprule
\tablehead{Proposal class} & \tablehead{Checker decision} \\
\midrule
Target-IR semantics & 200/200 rejected \\
Resource understatement & 80/80 rejected \\
Source semantics & 80/80 rejected \\
Immutable input binding & 40/40 rejected \\
Output-gate binding & 40/40 rejected \\
Platform binding & 40/40 rejected \\
Source--target mismatch & 20/20 rejected \\
\midrule
Adversarial total & 500/500 rejected \\
Benign configurations & 60/60 accepted \\
\bottomrule
\end{tabular}
\end{table}

\subsection{Extending the Relation Interface}
\label{sec:relation-onboarding}

The onboarding question is whether a relation withheld from the initial
checked registry can enter through the documented interface without changing
the core trust boundary.  The two extension studies examine complementary dimensions of interface reuse.
XLSX measures the code and integration needed to instantiate an existing
bounded-execution pattern.  Audit-log Top-$k$ tests a new bounded-state pattern
after the generic relation core is frozen.
Together, these studies distinguish extension along two axes: adding a new
library and source grammar for an existing execution pattern, and adding a new
retained-state pattern to the same relation contract.

\paragraph{Existing execution pattern.}
After freezing the initial four-relation code closure and its hashes, we
selected XLSX from a preregistered 12-family onboarding pool.  XLSX already
belonged to the six-family operator study but had not yet entered the checked
registry.  Onboarding added an independently implemented 216-nonblank-line
family checker, a 63-line bounded-VM operator, and a 197-nonblank-line
untrusted builder.  Common trusted infrastructure gained 12 nonblank wiring
lines: eight in checker routing, two in VM dispatch, and two in
registry/import plumbing.  The capacity lease, runtime, unique-selection
algorithm, and staged-publication logic were unchanged.  The substantive XLSX
semantics and resource arithmetic reside in its checker, VM operator, and
builder behind the common relation API.
\mbox{Almost} all onboarding code is relation-specific; the common
resource-control and publication machinery is reused.

The XLSX checker accepts all 12 legal window/output configurations, rejects all
64 rehashed semantic, resource, input-binding, and contract mutations, and is
the unique rule selected by automatic registry search.  On the frozen
100,000-row workbook, three fresh 128\mib cgroups all commit the same result:
physical peaks are 72.32--72.43\mib, below the independently recomputed
80.21\mib bound, with zero swap or memory-limit events.  The added relation
then passes through the existing checker, bounded VM, capacity lease, and
staged-publication path.
Appendix~\ref{app:relation-onboarding} records the separate protocol and
accounting for these onboarding-time canaries.

\begin{table}[t]
\centering
\caption{Relation-interface extension accounting.  Code-size counts are
nonblank lines; shared changes connect each plugin to the common runtime.}
\label{tab:relation-extension}
\small
\begingroup
\setlength{\tabcolsep}{3pt}
\renewcommand{\arraystretch}{1.08}
\renewcommand{\tabularxcolumn}[1]{m{#1}}
\begin{tabularx}{\columnwidth}{@{}
  >{\centering\arraybackslash}m{0.14\columnwidth}
  >{\centering\arraybackslash}m{0.43\columnwidth}
  >{\centering\arraybackslash}X@{}}
\toprule
\tablehead{Extension} & \tablehead{Relation-specific code} &
\tablehead{Shared-core changes} \\
\midrule
XLSX & \shortstack{Checker: 216\\VM operator: 63\\Builder: 197} &
\shortstack{12 wiring lines\\Core path unchanged} \\
\addlinespace[3pt]
Top-$k$ & \shortstack{Recognizer + facts\\Heap target + arena\\Postcondition} &
\shortstack{Registration + routing\\Dispatch + freeze\\7 core files unchanged} \\
\bottomrule
\end{tabularx}
\endgroup
\end{table}

\paragraph{New bounded-state pattern.}
We next froze the generic relation API and runtime before implementing an
audit-log Top-$k$ plugin.  The submitted program materializes 750,000 strict
JSONL events, globally sorts them by risk score with event-ID tie breaking,
and returns the first 64.  The canonical target scans once while retaining a
size-64 heap.  Seven generic-core files remain byte-identical.  Top-$k$ semantics
reside in the relation plugin; shared edits are limited to registration,
source routing, bounded-VM dispatch, and code-freeze plumbing.

The checker accepts 20/20 legal configurations and rejects every row in a
100-instance adversarial schedule, which contains 76 distinct certificate
hashes.  In three fresh 128\mib cgroups, checked execution commits 3/3 exact
results at 16.21--16.31\mib, below an 80.21\mib reconstructed bound.  The eager
global sort is OOM-killed in all three 128\mib runs, while the same source
completes 3/3 times with exact output at 2\,GiB and peaks at
772.33--772.38\mib.  This relation adds an explicit $O(k)$ retained-state
obligation to the four patterns represented by the first five plugins.
Appendix~\ref{app:relation-onboarding} gives the frozen protocol and result
identities.

\subsection{Enforcing Admission and Publication}
\label{sec:checked-admission}

We rerun one fixed source/input bundle for each relation in the frozen
five-relation admission configuration, using three fresh 128\mib cgroups per
bundle.  Each online container mounts only frozen runtime code, generated
source, immutable input, and its staged-output directory; the expected-output
artifact remains host-side.  All 15 executions pass their registered
online postcondition and resource gate, then commit; all record zero swap, OOM,
and memory-limit events and satisfy $P_{\mathrm{phys}}\leq U\leq B$.  Only
after each committed container exits does the host open the exact evaluation
oracle, obtaining 15/15 matches.  Median physical peaks are 14.71, 14.77,
14.67, 14.71, and 71.84\mib for CSV, FASTA, FCS, Zarr, and XLSX, respectively;
the corresponding three-repeat range widths are 0.07, 0.16, 0.13, 0.08, and
0.04\mib.  Checked bounds are 80.14--80.27\mib.  The shared deployment
manifest allocates a 64\mib runtime reserve and a 16\mib I/O reserve to every
relation; these fixed reserves dominate the bounds of the first four relations.
The present manifest prioritizes safe admission; relation-specific reserve
profiles would produce tighter host reservations.  Median fresh-cgroup
certification and execution phase wall times are 0.40--0.46 and
0.41--0.43\,s, respectively, for CSV, FASTA, FCS, and Zarr; the corresponding
XLSX times are 41.70 and 41.59\,s because both phases scan the workbook.
Certification is memory-bounded but is not always cheaper than execution.  For
XLSX, exact input-fact reconstruction currently requires a second full workbook
scan, so certification approximately doubles scan work.  This is a
relation-specific cost rather than a requirement of the shared runtime.

Appendix~\ref{app:physical} gives the per-relation checked bounds and measured
peak ranges for this same 15-transaction matrix.  The certifier evaluates every
relation registered in that frozen configuration and dispatches only a unique
accepted match.

\subsection{Capacity Safety and Failure Recovery}

Capacity leasing is relation-independent, so lease competition uses synthetic
24\mib allocations while lifecycle faults exercise the registered CSV
relation.  At 4, 8, and 16 contenders in fresh 256\mib cgroups,
across nine runs and 84 admission attempts, 36 acquire
one of four 24\mib leases and 48 are rejected before allocation.  Crashes are
injected at six lifecycle boundaries in each run, yielding 54 fault cases.
Every pre-publication crash exposes no output and releases the dead process's
lease after process-identity validation using the Linux boot ID and
\code{/proc/<pid>/stat} start time, preventing PID reuse from reclaiming a live
lease; all nine post-publication outputs remain verifier-valid.  No run records
swap or a memory-limit event.  This protocol isolates the lease and commit
invariants from operator-specific variation.

\begin{figure*}[t]
\centering
\includegraphics[width=\textwidth]{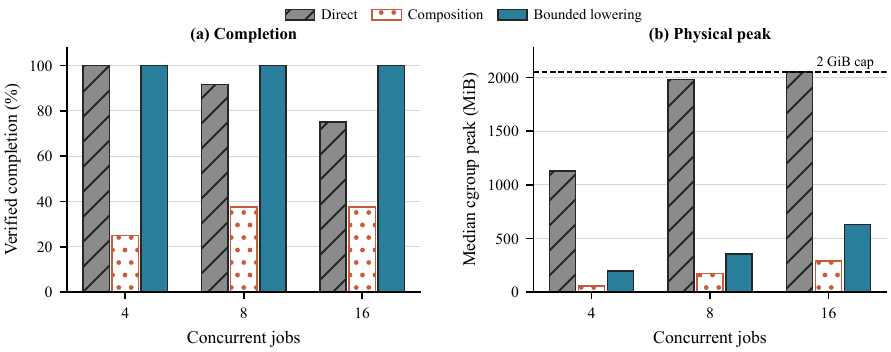}
\caption{Completion and measured cgroup peak under one 2\,GiB cgroup.
Panel~(a) shows verifier-passing completion; panel~(b) shows the corresponding
physical peak, with the dashed horizontal line marking the cap.  The hollow,
hatched composition bars reach only 37.5\%
completion at concurrency 8 and 16; its lower peak reflects less completed
work rather than greater capacity.  The bounded-lowering arm completes every submission while
remaining below the cap.}
\label{fig:concurrency}
\end{figure*}

Separately, we run the six-family operator-characterization lowerings in
family-balanced batches at 4, 8, and 16 concurrent jobs under one 2\,GiB
cgroup cap.  Figure~\ref{fig:concurrency} reports completion and measured cgroup
peak.  At concurrency 16, direct execution reaches the cap, completes
36/48 submissions, and yields 0.258 verified results/s.  Bounded lowering uses a
629.53\mib median peak, completes 48/48, and yields 0.462 verified results/s:
$1.79\times$ aggregate goodput.  Across the three repeats, bounded-lowering goodput is
0.446--0.473 verified results/s and peak is 620.87--647.01\mib; direct goodput
is 0.206--0.290/s and all three runs reach 2048\mib with OOM kills.  The
implemented composition has a small peak but rejects or produces incorrect
results.  Its smaller measured peak comes from executing only the 37.5\% of
submissions for which it has a valid candidate, not superior verified capacity.
Its observed rate at concurrency 8 is likewise inflated by the shortened,
mostly rejected batch.

An auxiliary uncapped semantic-throughput run on a 112-vCPU server separates
memory failure from operator cost.  At concurrency 16, both direct and bounded lowering
complete 48/48; bounded execution increases aggregate verified goodput from
0.461 to 0.543/s (17.9\%).  The main benefit is capacity safety, accompanied by
higher goodput in this control.  The matched latency study reports the same
direction over all 24 tasks.

\subsection{Integrating with Agent Runtimes}

The harness workload contains 16 tasks spanning CSV, FASTA, FCS, and Zarr under
a 24\mib configuration, and separates its tool boundary from the agent's final
response.
The native Qwen harness produces 16/16 schema-valid actions matching
the frozen tasks.  All request bounded execution, all 48 capped repeats verify
and commit, and all 16 final answers pass the declared key-and-type result
schema.  This arm validates identity and dispatch integration; the external
harness below exercises eager-plan lowering.

The external smolagents harness produces an actual framework tool call for all
16 tasks.  With no resource feedback, every call requests the eager access mode.
The registry lowers all 16 before dispatch; all 48 capped repeats verify and
commit, and replayed framework actions match their checked actions for 16/16
tasks.  Fifteen tasks issue exactly one call.  Strict final JSON verification is
13/16: two responses contain the expected values but violate the required JSON
format, and one repeats a checked Polars call three times before returning an
incorrect answer.  The result shows that verified
bounded tool execution composes with an external agent loop.  Tool-execution
validity and final-response policy are measured separately: the three
strict-final failures arise after tool execution from formatting or
repeated-call behavior.

The Mistral native replication generates matching bounded actions for all
16 tasks.  All 48 fresh-cgroup transactions verify, commit, and reproduce the
same tool result across repeats; 15/16 final answers match that result exactly.
The remaining Polars response changes a value while restating an already
verified committed result.  Together with the Qwen native arm, this separates
cross-model integration at the checked tool boundary from each model's
subsequent answer fidelity.

A deployment-owned bounded tool surface provides the strongest architectural
baseline.  Qwen emits matching typed actions for
16/16 native tasks and 15/16 smolagents tasks; Mistral emits matching actions
for 15/16 native tasks.  Every admitted action then completes all three fresh
24\mib runs: 48/48, 45/45, and 45/45, respectively.  The two planning failures
are retained (one Biopython call under Qwen/smolagents and one Zarr call under
Mistral).  Across these four harness families, a deployment that owns the
entire tool surface can obtain the same bounded physical execution with a
simpler typed API.  \system serves code-generating agents and legacy or
heterogeneous interfaces that expose both eager and bounded implementations;
in the unrestricted Qwen/smolagents path,
all 16 eager requests match a registered source relation and are lowered before
dispatch.

\section{Related Work}

\system sits at the intersection of three lines of work: compiling Skills and
agent traces into executable artifacts, transforming programs under resource
constraints, and enforcing budgets at Agent-runtime boundaries.  These lines
determine what can run, how an implementation can be improved, and where its
resource use is controlled.  We focus on the trust link between transformation
and dispatch, where a concrete model-generated program and immutable input meet
the physical capacity of one tool invocation.

\paragraph{Skill compilation and executable artifacts.}
SkillZip preserves procedural structure through section-graph compression,
dependency-closed hydration, verifier reachability, and reversible expansion
\cite{tan2026skillzip}; Control Under Compression studies the executable
reliability of compressed tool-control contexts \cite{hou2026control}.  Skill
compilers then make this structure operational in different ways.  SkVM
decomposes Skill requirements into capabilities, profiles model--harness
support, and solidifies or recompiles implementations for portable execution
\cite{chen2026skvm}.  SkCC introduces a typed Skill IR with compile-time
analysis \cite{ouyang2026skcc}; SkillSmith emits minimal executable interfaces,
policies, validation evidence, and fallback paths \cite{xu2026skillsmith}; and
SkillOpt uses trajectories and task verifiers to compile script-oriented Skill
artifacts \cite{rao2026skillopt}.  Related workflow compilers operate on plans
and traces: EvoC2F applies dependency- and effect-aware orchestration and
fault-tolerance transformations \cite{wei2026evoc2f}; TraceCompiler recovers
argument provenance and abstains when recovered effects are underdetermined
\cite{elyadouni2026tracecompiler}; Auto extracts guarded executable artifacts
from recorded behavior \cite{jaber2026auto}; and Profile--Then--Reason combines
explicit workflows, deterministic operators, trace verification, and bounded
repair \cite{enabe2026ptr}.

\paragraph{Checked transformation and resource reasoning.}
Dias dynamically applies precondition-checked rewrites to concrete Pandas
programs \cite{baziotis2024dias}; RuleScript provides a portable, verifiable
language for query-plan rewrites \cite{pan2026rulescript}; and RuleFlow turns
LLM-discovered Pandas optimizations into reusable compiler rules
\cite{singh2026ruleflow}.  Profile-guided systems discover and repair resource
problems at larger scope: MOA mines memory anti-patterns and produces static
checkers and codebase-scale patches \cite{liang2026moa}, PerfAgent iterates
repository patches using profiler and verifier feedback
\cite{deng2026perfagent}, and SWE-Pro benchmarks coding agents on
repository-level performance optimization, including parameterized peak-memory
tests \cite{sarikayak2026swepro}.  Classical work supplies complementary
foundations: resource-guided synthesis targets functional and symbolic resource
specifications \cite{knoth2019resyn}; program logics and automatic analyses
certify heap or whole-program bounds \cite{beringer2005heap,carbonneaux2015bounds};
and Ngo et al. study verification and synthesis of constant-resource
implementations for side-channel security \cite{ngo2017constant}.  JRes further
shows that runtime resource control can expose feedback for execution-plan
adaptation \cite{czajkowski1999jres}.

\paragraph{Budgeted Agent execution.}
CATP-LLM profiles tool costs and learns performance--cost-aware plans
\cite{wu2025catp}.  Agent Contracts formalizes task interfaces,
multidimensional budgets, success criteria, and budget-conserving delegation
\cite{ye2026contracts}.  Agent libOS integrates Agent processes, loaded Skills,
capabilities, budget preflight, and subprocess enforcement
\cite{zhang2026agentlibos}, while AgentCgroup characterizes tool-driven memory
spikes and controls tool-call cgroups with Agent intent and kernel mechanisms
\cite{zheng2026agentcgroup}.  Linux cgroup v2 provides the underlying memory
accounting and enforcement substrate \cite{linuxcgroupv2}.  Beyond memory,
PORTICO mediates revocable resource and effect capabilities before side effects
\cite{santos2026portico}, whereas AgentDoS documents resource-lifecycle
vulnerabilities in widely used agents \cite{luo2026agentdos}.

Together, prior work provides executable Skill representations, resource-aware
transformations, and runtime actuators.  \system connects them at dispatch: it
derives a candidate relation from the submitted program and immutable input,
independently rebuilds and checks the bounded target and its live-set bound,
then couples execution to an atomic capacity lease and postcondition-gated
publication.

\section{Limitations and Future Work}

The current implementation uses six relation plugins spanning five
bounded-execution patterns for deterministic, local, read-only tool programs
whose results can be checked before publication.

\paragraph{Postconditions.}
Each registered relation requires a publication check for its declared result.
The current implementation uses declarative or input-derived checks for
deterministic outputs.  Richer Agent tasks will require postconditions based on
schemas, invariants, relational constraints, or delayed validation.

\paragraph{Relation and source-form coverage.}
\system does not automatically verify arbitrary Python programs.  Each supported
computation requires an audited relation over a closed source grammar.  The
current implementation includes six relations spanning five bounded-execution
patterns.  Section~\ref{sec:relation-onboarding} shows that new relations can
reuse the common checker/runtime path, but relation-specific semantic auditing
remains required.  Broader Skill corpora and coding-Agent traces are needed to
measure coverage beyond these registered relations.

\paragraph{Platform calibration.}
Memory bounds are tied to a platform manifest covering the runtime, allocator,
page size, and fixed reserves.  Porting to another software or hardware stack
therefore requires fresh calibration and capped validation.  More granular
reserve profiles could reduce the conservatism of the unified manifest.

\paragraph{Stateful and remote effects.}
The current commit protocol stages local outputs.  Email, payment, and mutable
service calls require idempotency, authorization, and a transaction or
compensation protocol that couples the external effect to result publication.

\paragraph{Evaluation breadth.}
The executable baselines in this study share a common harness.  End-to-end
integration with released optimization systems, additional model families, and
larger multistep workloads will test interoperability and comparative
performance beyond the present evaluation.

\section{Conclusion}

Resource-sensitive Agent Skills package procedural knowledge for tool use.
Their physical consequences take shape when a model instantiates
that guidance as a concrete program over a concrete input.  Our study identifies
tool dispatch as the boundary where semantically valid plans can diverge
sharply in memory demand and hard-cap feasibility across heterogeneous
data-processing Skills.

\system places a checked trust boundary at that point.  For a program generated
from a routed Skill, a registered instance of the common five-obligation
relation contract identifies the admitted source computation, reconstructs its semantic
parameters and immutable input facts, validates a bounded target
configuration, and executes the checker-rebuilt target in a bounded VM under
capacity control.
Staged publication connects this physical decision to the declared task result,
while the planner contributes proposals rather than execution authority.
The system's generality therefore comes from reusing one trusted dispatch
architecture across audited relation plugins, rather than from accepting
arbitrary generated programs.

Operator characterization across six families shows that bounded access modes
preserve verified outputs and expand completion under external memory caps.
The registered plugins instantiate the same five-obligation contract across
streaming and row-window reductions, metadata projection, chunked reduction,
and bounded-state Top-$k$.  Across these instances, the checker and bounded VM
reject semantic and resource violations and complete isolated capped
transactions.  The XLSX onboarding study exercises the same core mechanisms
while measuring relation-specific implementation cost.
System integration studies exercise Agent harnesses, lease competition, and
failure recovery.  More broadly, \system turns registered resource-sensitive
relations for programs generated from routed Skills into enforceable physical
contracts at tool dispatch---a systems foundation for memory-bounded Agent
execution.

\FloatBarrier
\begingroup
\small
\interlinepenalty=10000
\bibliographystyle{unsrt}
\bibliography{refs}
\endgroup

\clearpage
\appendix
\counterwithin{figure}{section}
\counterwithin{table}{section}
\section*{\Large Appendix}

\section{Detailed Experimental Protocols}
\label{app:protocols}

\paragraph{Cap-feedback protocol.}
Qwen2.5-14B receives the exact family cap and input size on its first attempt.
After a failed physical run, it receives the previous code, process exit,
cgroup peak and OOM events, stderr/stdout and parse status, and one binary
verifier bit for at most two further attempts.  It never receives the expected
result, a lowering, or an API hint; success stops future rounds.  Each first
successful program is rerun three times in fresh cgroups.  These experiments
use the development caps; the externally fixed cap sweep remains the independent
cap-sensitivity study.

\paragraph{Natural plans.}
Qwen2.5-14B-Instruct-AWQ \cite{qwen2025qwen25} receives the naturally compressed
Skill and frozen task prompt.  Each of 24 tasks has three predeclared seeds,
temperature 0.2, and no verifier, memory, or lowering feedback during
generation.  We select the first verifier-valid plan in attempt order and only
then inspect its resource effects.  Mistral-7B receives the same tasks,
compressed Skills, seeds, and feedback policy; its four verifier-valid attempts
form a disclosed source-checker validation pool rather than an untouched test.

\paragraph{Harness protocols.}
The native and smolagents 1.26.0 harnesses reuse 16 tasks across the four
registered relations covered by this harness.  The native harness asks Qwen2.5-14B or Mistral-7B for
one typed action.  The smolagents harness retains its tool schemas, memory,
call processing, and final-answer loop; a transport-only adapter converts the
vLLM response into the framework's tool-call object.  Neither planner receives
a cap, verifier result, or lowering feedback.  For the bounded-only baseline,
the schema removes the access-mode choice and binds each tool identity directly
to its bounded handler.  Only task-matching actions enter three fresh 24\mib
runs; planning failures remain in the task denominator.

\paragraph{Five-relation admission and oracle isolation.}
The main physical-admission study freezes a 23-file online code closure,
container image, five registered relation schemas, and one source/input bundle
per relation.  CSV, FASTA, FCS, Zarr, and XLSX each run three times in fresh,
swap-disabled 128\mib cgroups.  Each online container mounts only the frozen
runtime code, generated source, immutable input, and staged-output directory;
the expected-output artifact remains host-side until the container exits.  The certifier and
checker identify the unique matching relation, reconstruct input facts, and
validate the target configuration and live set;
the runtime executes only that target in the bounded VM, requires both the
registered online postcondition and physical admission checks, and publishes
through a no-overwrite staged commit.  Only after the committed container exits
does the host open the exact evaluation oracle.  Per-cell records retain the
checked bound, cgroup peak, swap and memory events, commit state, and post-commit
oracle result.

\paragraph{Matched latency.}
At an external 2\,GiB cap, all 24 tasks run as one excluded warmup pair followed
by six measured adjacent AB/BA pairs.  Both arms use fresh cgroups and the same
verifier.  The measurement is warm-cache in-container launch-to-exit time.  It
excludes Docker create/start; the bounded arm includes concrete lowering,
envelope checking, tool execution, verification, and staged commit.

\section{Operator Characterization}
\label{app:operators}

\begin{table*}[t]
\centering
\caption{Bounded and eager access forms in each evaluated family.}
\label{tab:bounded-surfaces}
\small
\setlength{\tabcolsep}{4pt}
\begin{tabular}{@{}
  >{\centering\arraybackslash}m{0.12\textwidth}
  >{\centering\arraybackslash}m{0.40\textwidth}
  >{\centering\arraybackslash}m{0.40\textwidth}@{}}
\toprule
\tablehead{Family} & \tablehead{Bounded form} & \tablehead{Eager form} \\
\midrule
XLSX & \code{load\_workbook(read\_only=True)} or projected
\code{read\_excel(usecols=...)} & The same loaders without the mode or column
projection \\
FlowIO & \code{FlowData(..., only\_text=True)} & The same constructor without
\code{only\_text}, materializing event data \\
AnnData & \code{read\_h5ad(..., backed='r')} & The same reader without
\code{backed}, materializing the matrix \\
Zarr & Chunked slices and an incremental reduction & \code{array[:]} on the
same array object \\
Polars & \code{scan\_csv(...).collect(engine='streaming')} & Sibling
\code{read\_csv} followed by the same relational operations \\
Biopython & Single-pass consumption of the \code{SeqIO.parse} iterator & The
caller wraps the same iterator in \code{list(...)} \\
\bottomrule
\end{tabular}
\end{table*}

\begin{figure}[H]
\centering
\includegraphics[width=\columnwidth]{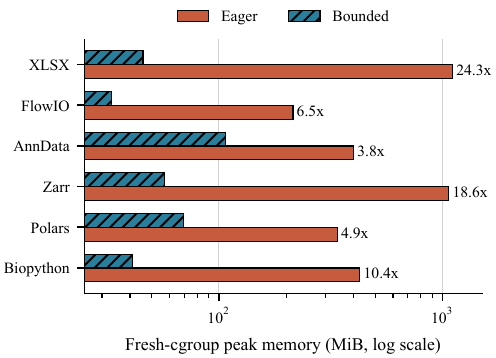}
\caption{Largest-input eager and bounded physical peaks.  Ratios are annotated
at the eager bars; paired executions produce the same verified result.}
\label{fig:memory}
\end{figure}

All 36 family--scale--mode cells pass their verifier in every repeat.  At the
largest scale, Figure~\ref{fig:memory} shows that bounded implementations reduce
fresh-cgroup peaks by $3.75$--$24.25\times$ (median $8.45\times$).  The spread
explains why access mode belongs in the resource contract: a single reserve
based on input size is wasteful for metadata and streaming tasks but unsafe for
eager arrays and workbooks.

\FloatBarrier
\raggedbottom
\section{Development-Cap Prompting and Repair}
\label{app:development}

Table~\ref{tab:hardcap} separates construction, semantic validity, and physical
failure under the per-family development caps.  These caps exercise mechanism
behavior; the externally fixed cap sweep in Section~\ref{sec:external-cap}
provides the cross-cap completion result.  Always pinning the resource clause yields a
correct bounded program for 9/24 tasks, and ordinary retry for 4/24.  Thus the
latency benefit of bounded operators does not imply that a prompt reliably
constructs one.

\begin{table}[H]
\centering
\caption{Constructive outcomes at per-family development caps in operator
characterization.  ``Passed'' counts verifier-passing runs; failures are
reported as cap/verifier/abstention counts.}
\label{tab:hardcap}
\small
\begin{tabular}{@{}cccc@{}}
\toprule
\tablehead{Method} & \tablehead{Tasks} & \tablehead{Passed} & \tablehead{Failures} \\
\midrule
Direct eager & 0/24 & 0/72 & 72/0/0 \\
Always-pin & 9/24 & 27/72 & 12/33/0 \\
Retry & 4/24 & 12/72 & 9/51/0 \\
Composition & 9/24 & 27/72 & 0/15/30 \\
\midrule
Bounded lowering & \textbf{24/24} & \textbf{72/72} & \textbf{0/0/0} \\
\bottomrule
\end{tabular}
\end{table}

The profile gate and plan gate are reject-only controls: each abstains on all
72 runs and therefore incurs no physical or verifier failure.  They quantify
safe non-construction as a separate control class.

The cap-feedback baseline completes 5/24 tasks after 67 total attempts: two in
round one, one in round two, and two in round three.  All five selected programs
then pass three fresh repeats.  The successes are three Biopython and two
FlowIO tasks; AnnData, Polars, XLSX, and Zarr remain 0/4.  Physical feedback is
therefore useful for a minority of calls but does not replace deterministic
lowering in this matrix.

\FloatBarrier
\section{Frontend, Scale Transfer, and Latency}
\label{app:coverage}

On 20 held-out positive Skills, the static frontend accepts 13 dispatchable
contracts; effect-kind and evidence are recovered for 14/20.  All five
negative controls abstain.  The six-family envelope is calibrated on four
input scales and evaluated on a fifth.  None of six admission-eligible bounded
cells exceeds its bound; one eager estimate is ineligible for small-cap
dispatch.  Median and maximum reserve overhead among admitted cells are
44.7\% and 99.3\%.  Across 10,000 preflight iterations, AST capture plus
envelope instantiation takes 0.804 ms median and 1.260 ms P95.

All 144 measured latency pairs are valid.  The median bounded/direct ratio over
24 task-level medians is 0.811, and every task is below one (range
0.590--0.958).  Median bounded preflight and checker phases are 2.744 and
0.839 ms; the operator dominates the measured pipeline.  Faster bounded
operators do not make prompting a reliable enforcement mechanism: the
always-pin baseline constructs a verifier-passing bounded program for only
9/24 tasks in Table~\ref{tab:hardcap}.  Table~\ref{tab:latency} reports family
medians.

\begin{table}[H]
\centering
\caption{Matched warm-cache in-container process wall time at 2\,GiB.  Each row
reports the median time over four tasks, each measured with six AB/BA pairs.
Ratio is the median of the four per-task ratios, not the ratio of the two time
medians.}
\label{tab:latency}
\small
\begin{tabular}{@{}cccc@{}}
\toprule
\tablehead{Family} & \tablehead{Direct (s)} & \tablehead{Bounded (s)} & \tablehead{Ratio} \\
\midrule
XLSX & 24.381 & 19.727 & 0.811 \\
FlowIO & 0.534 & 0.449 & 0.833 \\
AnnData & 1.355 & 1.098 & 0.846 \\
Zarr & 1.220 & 1.057 & 0.879 \\
Polars & 1.069 & 0.787 & 0.740 \\
Biopython & 3.136 & 1.947 & 0.622 \\
\bottomrule
\end{tabular}
\end{table}

\FloatBarrier
\section{Five-Relation Physical Admission}
\label{app:physical}

Table~\ref{tab:v2cert} gives the per-relation physical values behind the unified
15-transaction study in Section~\ref{sec:checked-admission}.  Each row
summarizes three fresh cgroups under the same 128\mib cap and frozen
five-relation admission configuration.  Every run passes its online postcondition, commits
before host-side exact evaluation, records zero swap, OOM, and memory-limit
events, and satisfies $P_{\mathrm{phys}}\leq U\leq B$.

\begin{table}[H]
\centering
\caption{Per-relation physical validation of the frozen five-relation admission
configuration.  Memory values are MiB over three fresh 128\mib cgroups per relation;
cert./exec. reports median capped phase wall times in seconds.}
\label{tab:v2cert}
\small
\resizebox{\columnwidth}{!}{%
\begin{tabular}{@{}ccccc@{}}
\toprule
\tablehead{Relation} & \tablehead{Bound} & \tablehead{Median peak} &
\tablehead{Peak range} & \tablehead{Cert./exec. (s)} \\
\midrule
CSV / Polars & 80.142 & 14.707 & 14.672--14.746 & 0.410/0.418 \\
FASTA / Biopython & 80.142 & 14.770 & 14.641--14.801 & 0.456/0.432 \\
FCS / FlowIO & 80.143 & 14.668 & 14.652--14.785 & 0.398/0.412 \\
Zarr & 80.266 & 14.711 & 14.664--14.746 & 0.409/0.426 \\
XLSX & 80.209 & 71.844 & 71.801--71.844 & 41.700/41.587 \\
\bottomrule
\end{tabular}
}
\end{table}

\FloatBarrier
\section{Relation-Extension Protocols}
\label{app:relation-onboarding}

The XLSX onboarding study in Section~\ref{sec:relation-onboarding} uses a
separately frozen code closure, mutation suite, workbook, and cgroup protocol.
These 64 onboarding-time mutation canaries are separate from the
five-relation validation matrix in Section~\ref{sec:checker-eval}, which regenerates 100
mutations per relation (500 total).  The XLSX physical repeats are likewise
separate from the uniform five-relation admission matrix in
Section~\ref{sec:checked-admission}; the two frozen configurations yield the
72.32--72.43\mib onboarding range and the 71.84\mib unified-profile median,
respectively.

The Top-$k$ semantic-breadth protocol is frozen separately before plugin
implementation.  It fixes a 750,000-record, 124,455,560-byte strict JSONL
ledger, $k=64$, the ordering \code{risk\_score DESC, event\_id ASC}, a
128\mib primary cap, and a 2\,GiB eager-validation cap.  The primary schedule
uses three balanced eager/checked repeats; the 2\,GiB control uses three eager
repeats.  The evaluation output remains host-side and is opened only after
container exit.  The checker schedule contains 20 legal
configurations and 100 adversarial rows (76 distinct certificate hashes).
The generic-core freeze, checker summary, and physical summary are separately
hash-bound; no Top-$k$ result is pooled into the five-relation matrices above.

\end{document}